\documentclass[sigconf]{acmart}
\usepackage{booktabs}
\usepackage{multirow}
\usepackage{graphicx}
\AtBeginDocument{%
  }

\setcopyright{acmlicensed}
\copyrightyear{2018}
\acmYear{2018}
\acmDOI{XXXXXXX.XXXXXXX}
\acmConference[Conference acronym 'XX]{Make sure to enter the correct
  conference title from your rights confirmation email}{June 03--05,
  2018}{Woodstock, NY}
\acmISBN{978-1-4503-XXXX-X/2018/06}

\newcommand{\ours}{\textsc{{SecureGARD}}}
\begin{document}

\title{SecureGARD: Dynamic Ontology-Guided Governance \\ for  Secure Tool-Using LLM Agents}

\author{Guoqin Xiong}
\authornote{Both authors contributed equally to this work.}
\affiliation{%
  \department{MIIT Key Laboratory of Pattern Analysis and Machine Intelligence, College of Computer Science and Technology}
  \institution{Nanjing University of Aeronautics and Astronautics}
  \city{Nanjing}
  \country{China}
}
\email{guoqin\_xiong@nuaa.edu.cn}

\author{Shengtao Wen}
\authornotemark[1]
\affiliation{%
  \department{MIIT Key Laboratory of Pattern Analysis and Machine Intelligence, College of Computer Science and Technology}
  \institution{Nanjing University of Aeronautics and Astronautics}
  \city{Nanjing}
  \country{China}
}
\email{shengtao\_wen@nuaa.edu.cn}

\author{Xiang Chen}
\authornote{Corresponding author.}
\affiliation{%
  \department{MIIT Key Laboratory of Pattern Analysis and Machine Intelligence, College of Computer Science and Technology}
  \institution{Nanjing University of Aeronautics and Astronautics}
  \city{Nanjing}
  \country{China}
}
\email{xiang\_chen@nuaa.edu.cn}

\author{Xinyi Zeng}
\affiliation{%
  \institution{Tsinghua University}
  \city{Beijing}
  \country{China}
}

\author{Zhaoxin Fan}
\affiliation{%
  \institution{Beihang University}
  \city{Beijing}
  \country{China}
}

\author{Wei Ye}
\affiliation{%
  \department{College of Electronic and Information Engineering}
  \institution{Tongji University}
  \city{Shanghai}
  \country{China}
}

\author{Xiaofeng Cao}
\affiliation{%
  \department{School of Computer Science and Technology}
  \institution{Tongji University}
  \city{Shanghai}
  \country{China}
}

\author{Sheng-Jun Huang}
\affiliation{%
  \department{MIIT Key Laboratory of Pattern Analysis and Machine Intelligence, College of Computer Science and Technology}
  \institution{Nanjing University of Aeronautics and Astronautics}
  \city{Nanjing}
  \country{China}
}
\email{huangsj@nuaa.edu.cn}

\renewcommand{\shortauthors}{Xiong et al.}


\begin{abstract}
Tool-using large language model (LLM) agents translate model outputs into consequential actions, turning indirect prompt injection into an execution-security problem. Existing defenses constrain tool calls, trajectories, or information flows, but often leave action semantics, state dependencies, and evidence relations implicit, limiting generalization across heterogeneous tools and complicating safe updates. We present \ours{}, a dynamic ontology-guided governance framework. \ours{} grounds heterogeneous calls in a shared semantic representation and combines an executable ontology with episode-level task state to derive deterministic ready, waiting, and forbidden decisions. A waiting decision identifies an unmet but recoverable obligation, enabling bounded replanning without expanding the agent's authorization boundary. \ours{} evolves its ontology offline from verified training executions. An offline LLM proposes instance-independent action--state axioms, which are accepted only if they address ontology errors exposed by training failures, preserve logical consistency and protected entailments, and pass held-out validation before the ontology is frozen. We evaluate \ours{} on AgentDojo, ASB, and AgentDyn under benign and indirect prompt-injection settings. \ours{} achieves a 0\% attack success rate (ASR) across all three benchmarks while improving task utility under attack compared with existing baselines. These results show that explicit semantic governance can strengthen execution security without indiscriminate blocking.
\end{abstract}




\maketitle


\section{Introduction}
\label{sec:introduction}

Large language models (LLMs) are increasingly deployed as agents that plan, invoke external tools, and interact with dynamic environments to accomplish user goals~\cite{yao2023react,schick2023toolformer,li2023apibank,qin2024toolllm}. Tool access extends LLMs from response generation to consequential task execution, but also exposes agent behavior to untrusted environmental content. In particular, indirect prompt injections embedded in emails, webpages, documents, or tool outputs may be interpreted as executable instructions, redirecting tool selection, altering argument bindings, or inducing actions that conflict with the original user request~\cite{abdelnabi2023indirect,zhan2024injecagent,debenedetti2024agentdojo}. Recent benchmarks demonstrate that these threats persist across heterogeneous applications, attack strategies, and dynamically evolving tool environments~\cite{debenedetti2024agentdojo,zhang2025asb,xiang2026agentdyn}. These risks also arise in Web-facing applications, making the governance of tool-mediated execution a core security problem for LLM agents with delegated user authority.

Existing defenses constrain agent behavior through instruction--data separation, injection detection, information-flow control, privilege policies, runtime specifications, or trajectory validation~\cite{hines2024spotlighting,chen2024struq,debenedetti2025camel,shi2025progent,wang2025agentspec,li2025drift}. Although these mechanisms provide important protection, they primarily operate over prompts, concrete tool interfaces, data flows, or planned execution sequences. A tool call exposes an interface name and a set of arguments, but does not necessarily represent the capability being invoked, the states on which it depends, the effects it may produce, or the evidence required to establish those states. Consequently, semantically equivalent tools may require separately defined constraints, execution failures may not reveal the underlying state dependencies, and newly identified governance knowledge may remain tied to a particular interface or trajectory.

\begin{figure}[!t]
    \centering
    \includegraphics[width=\columnwidth]{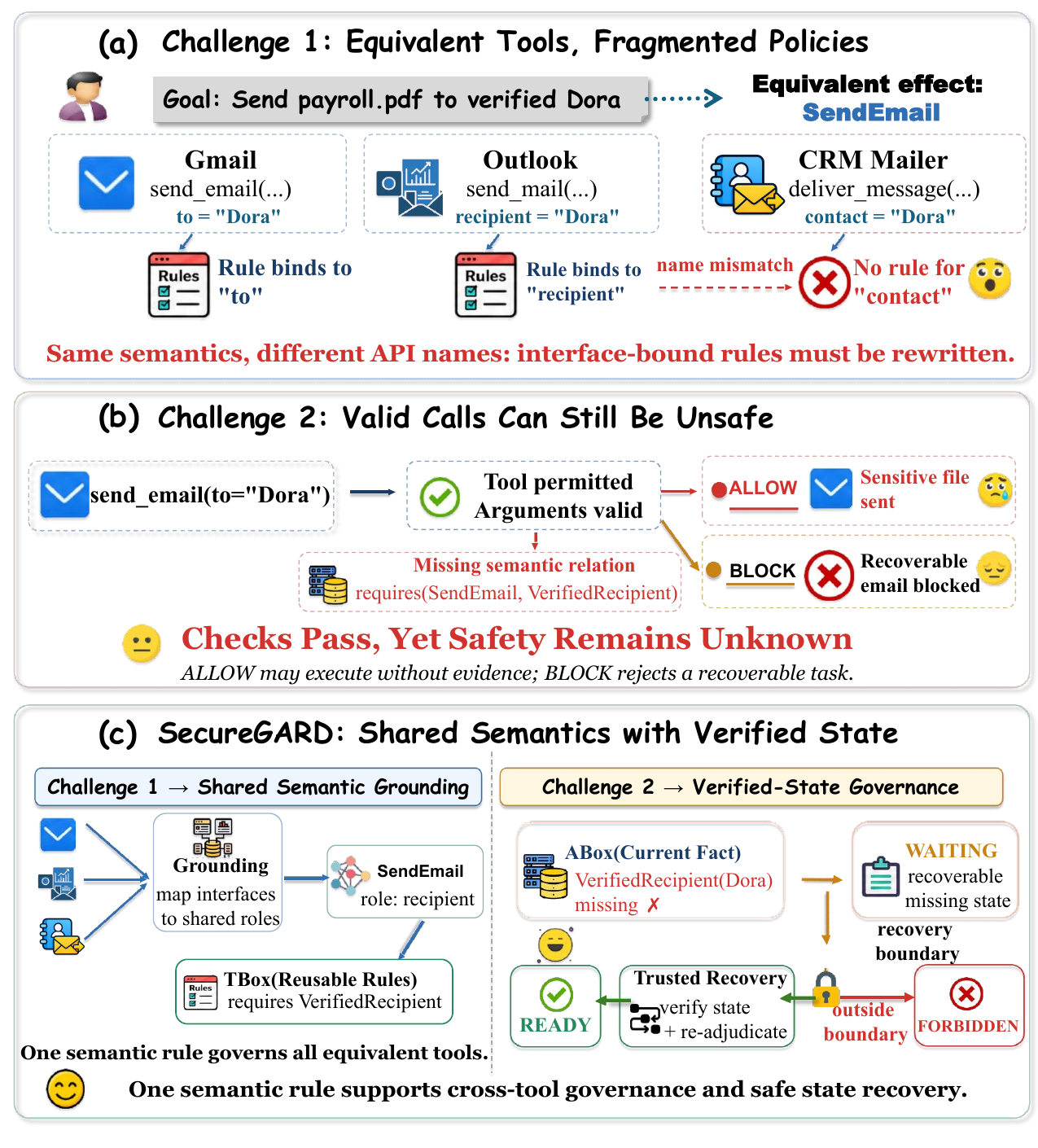}
    \caption{Motivation for \ours{}. (a) Interface-specific policies fragment equivalent tool actions. (b) Permission and argument checks cannot verify semantic prerequisites, making binary admission unsafe or overly restrictive. (c) \ours{} grounds heterogeneous calls into shared actions and checks TBox requirements against verified ABox facts, enabling three-way adjudication and bounded recovery.}
    \label{fig:motivation}
    \Description{Three vertically arranged panels illustrate the motivation for SecureGARD. The first shows equivalent email tools with different interfaces, preventing interface-specific policies from transferring. The second shows that permission and argument checks may either allow an unsupported action or block a recoverable task. The third shows SecureGARD grounding heterogeneous tools into a shared semantic action, checking TBox requirements against verified ABox facts, and returning ready, waiting, or forbidden decisions.}
    \vspace{-8pt}
\end{figure}

These limitations expose two complementary gaps, summarized in Figure~\ref{fig:motivation}. First, interface-specific policies do not readily transfer across tools that realize the same semantic action (Figure~\ref{fig:motivation}(a)). Second, permission and argument checks cannot determine whether the state prerequisites of an otherwise valid action have been established, forcing a binary defense to either admit an unsupported action or reject a recoverable task (Figure~\ref{fig:motivation}(b)). \ours{} addresses both gaps by grounding heterogeneous tool calls into shared semantic actions and adjudicating them against explicit action--state dependencies (Figure~\ref{fig:motivation}(c)). We represent reusable schema-level requirements and effects in the \emph{terminological component} (\textbf{TBox}), while verified facts about the current execution are maintained in the \emph{assertional component} (\textbf{ABox}).

Ontologies provide a suitable foundation for this representation because they encode concepts, relations, and constraints in a machine-interpretable form~\cite{hogan2021knowledge,pan2024unifying}. In tool-using agents, an execution ontology can abstract actions from their concrete implementations and connect them to environmental states through explicit semantic relations. An action may require a state before execution, establish or invalidate a state after successful execution, or observe a state through structured evidence. The resulting ontology therefore serves as an executable semantic layer for runtime governance rather than a passive description of domain knowledge.

We present \ours{}, a dynamic ontology-guided governance framework for secure tool-using LLM agents. \ours{} grounds heterogeneous tool calls into semantic action instances that capture their action classes, effects, parameter roles, and execution contexts. During execution, the frozen TBox supplies reusable action--state rules, while the episode-specific ABox records verified facts about the current environment. A separate episode-level TaskState records the procedural progress of the current task. Ontology reasoning determines whether an action's semantic requirements are satisfied, whereas TaskState preserves the execution context required for multi-step coordination and bounded replanning.

Based on the execution boundary and inferred action--state requirements, \ours{} deterministically classifies each grounded action as \emph{ready}, \emph{waiting}, or \emph{forbidden}. A \emph{ready} action satisfies its semantic requirements and may proceed. A \emph{waiting} action has an unmet but recoverable requirement; \ours{} records the missing obligation in TaskState and permits bounded recovery without expanding the authorization boundary. A \emph{forbidden} action violates the execution boundary and is rejected. Every initial or replanned action remains subject to the same semantic grounding and runtime adjudication procedure.

To accommodate previously unmodeled action effects and state dependencies, \ours{} evolves its ontology through a controlled offline process. Verified training executions are attributed to corresponding action--state changes, while outcomes caused by planning errors, tool failures, or unreliable observations are excluded from ontology learning. An offline LLM proposes instance-independent action--state axioms from bounded summaries of eligible evidence, but cannot directly modify the ontology or participate in runtime adjudication. Candidate revisions must be supported by independent training-task evidence and pass logical-consistency, semantic-preservation, and held-out validation checks. The accepted TBox is frozen before evaluation and deployment, while online executions may update only the episode-specific ABox and TaskState.

We evaluate \ours{} on AgentDojo, ASB, and AgentDyn under benign and indirect prompt-injection settings~\cite{debenedetti2024agentdojo,zhang2025asb,xiang2026agentdyn}. The ontology is evolved exclusively from the AgentDojo training split, then frozen and transferred to ASB and AgentDyn without target-benchmark revision. Across all evaluated tasks and backbones, \ours{} achieves an observed attack success rate (ASR) of 0\% while maintaining competitive task utility under both benign and attacked executions. Our contributions are summarized as follows:

\begin{itemize}
    \item We introduce \ours{}, an ontology-guided framework that grounds heterogeneous tool calls into shared semantic actions and supports \emph{ready}, \emph{waiting}, and \emph{forbidden} decisions over verified execution states.
    \item We develop an evidence-grounded ontology evolution procedure that derives and validates reusable action--state axioms from verified training trajectories before freezing the ontology for deployment.
    \item Experiments on AgentDojo, ASB, and AgentDyn demonstrate consistent security across benchmarks and backbones while maintaining competitive task utility.
\end{itemize}

\section{Related Work}
\label{sec:related_work}

\subsection{LLM Agents and Tool Use}

Large language models (LLMs) have emerged as key components of autonomous agents, providing reasoning and decision-making capabilities in complex environments. ReAct introduced an influential paradigm that interleaves reasoning and acting within a unified execution trajectory, while Toolformer demonstrated that language models can learn when and how to invoke external tools, extending LLMs beyond text generation toward interactive environments~\cite{yao2023react,schick2023toolformer}. Subsequent work has expanded this paradigm to large and heterogeneous tool collections. API-Bank provides an executable benchmark for tool-augmented LLMs, while ToolLLM develops instruction-tuning and retrieval mechanisms for mastering a broad range of real-world APIs~\cite{li2023apibank,qin2024toolllm}.

Beyond individual tool calls, effective agents must coordinate multi-step plans and retain information across extended interactions. Planning-oriented approaches decompose complex objectives into intermediate actions, whereas memory architectures such as MemGPT maintain and retrieve relevant interaction history over long-running tasks~\cite{huang2024llmagentplanning,packer2023memgpt}. Complementary work on efficient reasoning, such as CEEH, uses difficulty-aware entropy regularization to compress reasoning on easier questions while preserving exploration on harder ones~\cite{luo2026compress}. These advances broaden agents' reasoning and interaction capabilities, but do not by themselves establish whether a proposed external action is authorized or supported by the current environment.

\subsection{Defense and Governance for Tool-Using Agents}

Reliable agent execution requires controlling not only final responses but also intermediate decisions, tool interactions, and execution states. Existing approaches improve reliability through model-level defenses and external runtime governance. Model-level methods strengthen the separation between instructions and data or introduce dedicated detectors for unsafe or injected content, as exemplified by StruQ, Llama Guard, and InjecGuard~\cite{chen2024struq,inan2023llamaguard,liu2025injecguard}. Related reliability research includes DeCo, which uses intermediate-layer information to correct multimodal decoding, and UNIHD, which employs auxiliary tools to verify generated claims~\cite{wang2025deco,chen2024unified}. Their focus on content reliability complements execution governance, which additionally checks action authorization and environmental prerequisites.

Recent system-level approaches regulate agent execution beyond the model boundary. CaMeL separates trusted control logic from untrusted information flows and tracks their dependencies, while Progent, AgentSpec, and DRIFT enforce tool-use constraints through structured policies, runtime specifications, and execution validation mechanisms~\cite{debenedetti2025camel,shi2025progent,wang2025agentspec,li2025drift}. Although these approaches improve the reliability of agent execution, their control knowledge is mainly expressed through tool interfaces, execution trajectories, or predefined policies. Such representations lack a unified semantic abstraction for capturing the meaning of actions, their effects, and the dependencies between actions and environmental states, making governance knowledge difficult to generalize across heterogeneous tool environments.

\subsection{Ontology-based Semantic Modeling}

Ontologies and knowledge graphs provide structured representations of concepts, relations, and constraints, enabling machine-interpretable modeling beyond the implicit knowledge encoded in language models~\cite{hogan2021knowledge,pan2024unifying}. KnowPrompt incorporates relation-label semantics and structured constraints into prompt tuning for relation extraction, while MKGformer uses multi-level textual and visual fusion for multimodal knowledge graph completion~\cite{chen2022knowprompt,chen2022hybrid}. These methods illustrate how relational structure and heterogeneous evidence can support knowledge learning. Recent studies further integrate structured semantics with LLMs to improve knowledge grounding and reasoning. LLMs4OL explores LLM-assisted ontology construction, while OaK introduces a task-oriented ontology kernel that combines a schema, knowledge graph, and typed reasoning functions~\cite{babaei2023llms4ol,zhang2026oak}.

These approaches demonstrate the value of explicit semantic structures for agent reasoning and task understanding. However, existing ontology-enhanced agents mainly use semantic representations for knowledge organization, rather than for governing security-sensitive actions or validating execution decisions. Moreover, static semantic models cannot readily accommodate newly introduced tools, action effects, and state dependencies in open agent environments. Research on principled knowledge editing highlights that updates must account for knowledge interdependence and reasoning consistency~\cite{zhang2026principled}, a concern also relevant to revising reusable action--state rules. Adaptive agents such as Reflexion and ExpeL improve future behavior by incorporating feedback from previous executions~\cite{shinn2023reflexion,zhao2023expel}, but provide no explicit guarantee that such updates preserve previously validated behavior. \ours{} builds upon these directions by introducing a dynamic ontology as a semantic control layer for tool-using agents, enabling adaptive governance while preserving established behavioral constraints.

\section{Method}
\label{sec:method}

\ours{} is a dynamic ontology-guided governance framework for tool-using LLM agents. It introduces an explicit semantic layer between agent planning and external tool execution, where tool calls are grounded into semantic actions and governed through an evolving ontology. By decoupling action semantics from concrete tool implementations, \ours{} enables transferable and adaptive execution governance.

\begin{figure*}[t]
    \centering
    \includegraphics[width=\textwidth]{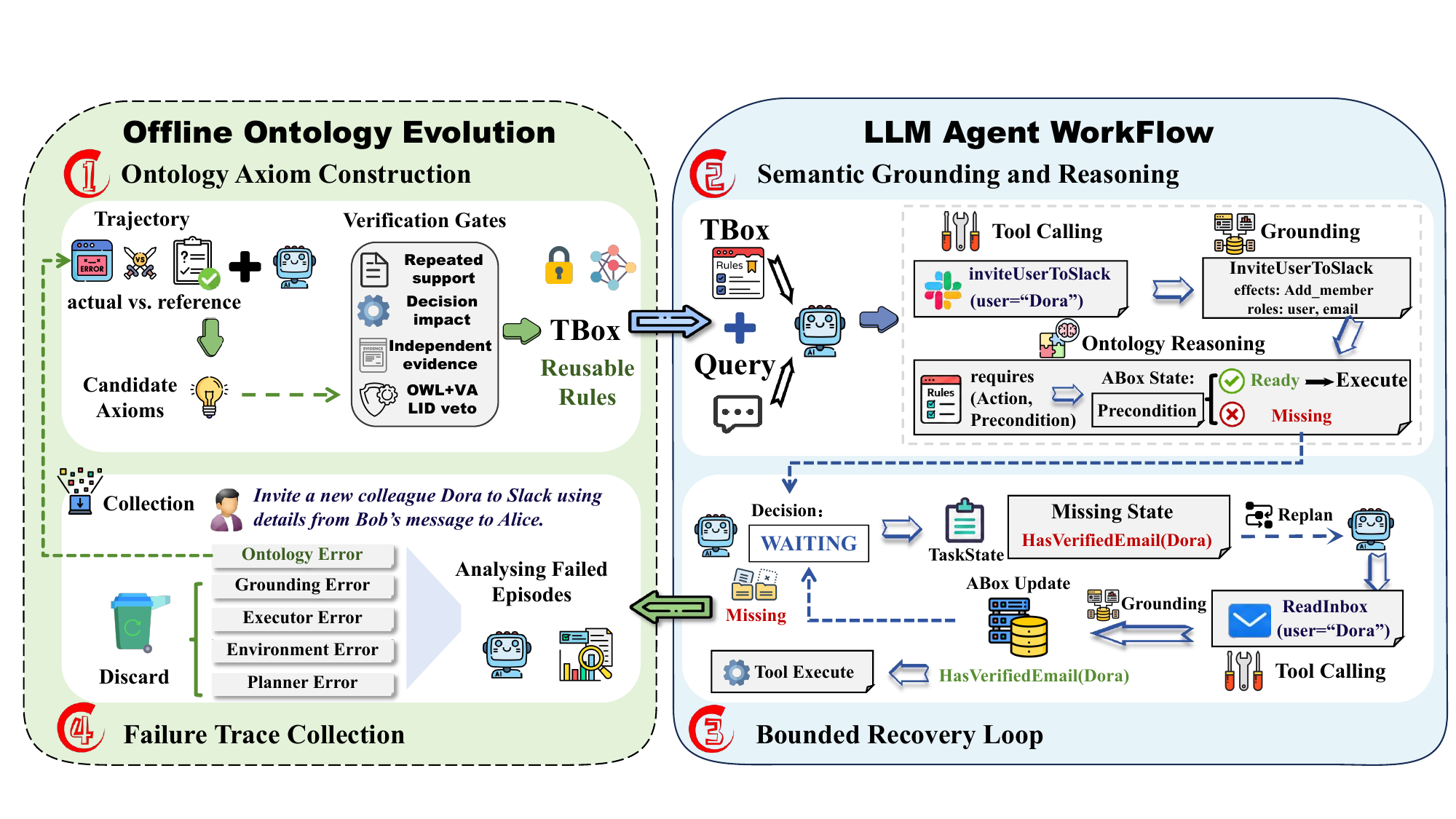}
    \caption{Overview of \ours{}. (1) Verified training trajectories generate candidate action--state axioms and reusable TBox rules. (2) Online tool calls are grounded and checked against current ABox facts. (3) Missing but recoverable requirements produce a \emph{waiting} decision, are recorded in TaskState, and trigger bounded recovery before re-adjudication as \emph{ready}. (4) Failure traces are filtered so that only ontology-relevant evidence supports subsequent offline evolution.}
    \label{fig:securegard_overview}
\end{figure*}

The framework consists of three components: semantic execution modeling, ontology-guided runtime governance, and evidence-grounded ontology evolution. During execution, \ours{} maintains an execution ontology and an episode-level TaskState for semantic reasoning and task coordination. As illustrated in Figure~\ref{fig:securegard_overview}, verified training evidence is attributed and converted into candidate action--state axioms through an offline revision process, while runtime tool calls are grounded, adjudicated, and executed against the resulting semantic knowledge. Ontology evolution is performed offline, and the accepted TBox is frozen before evaluation and deployment.

\subsection{Semantic Execution Modeling}
\label{sec:semantic_modeling}

Tool-using agents interact with heterogeneous environments through diverse interfaces. Different tools may expose distinct names, argument schemas, and invocation formats while realizing semantically equivalent operations. Defining execution constraints directly over tool interfaces therefore limits their transferability across environments. \ours{} introduces a semantic execution representation that separates the meaning of an action from its concrete implementation. Given a planner-generated tool call

\begin{equation}
    c_t=(u_t,\theta_t),
\end{equation}where $u_t$ denotes the concrete tool interface selected by the planner and $\theta_t$ represents the associated parameter bindings required for invocation, the grounding process constructs a semantic action instance

\begin{equation}
    x_t=
    \left\langle
        a_t,\mathcal{E}_t,\mathcal{R}_t,\mathcal{C}_t
    \right\rangle ,
\end{equation}where $a_t$ denotes the action class, $\mathcal{E}_t$ represents its semantic effects, $\mathcal{R}_t$ captures the semantic roles of its parameters, and $\mathcal{C}_t$ describes the execution context, including plan membership, tool source, capability status, parameter provenance, verified evidence, and the current episode-level TaskState. This representation establishes a shared semantic interface between heterogeneous tools and the governance layer. Instead of defining execution constraints over individual tool implementations, \ours{} reasons over abstract action semantics and their associated effects, roles, and dependencies. Therefore, governance knowledge can transfer across tools that realize equivalent functional capabilities.

\subsection{Ontology-Guided Runtime Governance}
\label{sec:runtime_governance}

\paragraph{\textbf{Semantic State Representation.} }\ours{} maintains two complementary representations during execution: semantic execution knowledge and procedural execution state. Semantic knowledge is represented by an ontology that captures relations among actions, environmental states, and execution contexts. At execution step $t$, the ontology is represented as

\begin{equation}
    \mathcal{O}_t=(\mathcal{T},\mathcal{A}_t),
\end{equation}where $\mathcal{T}$ is the cross-task TBox containing reusable semantic knowledge over actions and states, and $\mathcal{A}_t$ is the episode-specific ABox containing verified runtime facts. This separation prevents instance-level facts from directly becoming reusable knowledge. To support multi-step execution and recovery, \ours{} additionally maintains an episode-level TaskState

\begin{equation}
    \mathcal{Q}_t=
    \left\langle
        \mathcal{H}_t,\mathcal{V}_t,
        \mathcal{P}_t,\mathcal{B}_t
    \right\rangle ,
\end{equation}where $\mathcal{H}_t$ records completed action instances, $\mathcal{V}_t$ is a compact projection of verified facts exposed to the planner, $\mathcal{P}_t$ contains pending action instances, and $\mathcal{B}_t$ records unresolved obligations and their bounded recovery context. The ABox represents verified facts about the current environment, whereas TaskState tracks task progress and supplies structured context for bounded replanning. TaskState thus serves as an execution coordination state rather than general-purpose memory: it excludes planner reasoning and untrusted free-form content, cannot independently authorize actions, and remains subject to semantic grounding and runtime adjudication.

For an action class $a$ and a state class $s$, \ours{} defines four action--state relations:

\begin{equation}
\begin{aligned}
    &\mathsf{requires}(a,s), &&
    \mathsf{establishes}(a,s),\\
    &\mathsf{invalidates}(a,s), &&
    \mathsf{observes}(a,s).
\end{aligned}
\label{eq:ontology_relations}
\end{equation}The four relations define the operational semantics of action execution. $\mathsf{requires}$ specifies the verified states that must hold before an action can proceed; $\mathsf{establishes}$ and $\mathsf{invalidates}$ determine how successful execution updates the ABox; and $\mathsf{observes}$ identifies states that can be verified from structured observations. They therefore support pre-execution condition checking and post-execution state updates. Through action-class inheritance, constraints defined over an abstract capability apply uniformly to heterogeneous tools grounded to that capability.

\paragraph{\textbf{Execution Boundary and Semantic Adjudication.} }To prevent ontology evolution from expanding agent authority, \ours{} separates the execution boundary from revisable action--state knowledge. The execution boundary captures authority-defining constraints derived from the user request, tool capabilities, parameter provenance, and security-relevant bindings. Ontology evolution may refine semantic dependencies within this boundary but cannot enlarge the agent's authorized action space. The initial TBox includes one fixed seed axiom that forbids state-changing actions not grounded in the original user goal or an admitted plan revision. For an action satisfying the execution boundary, \ours{} evaluates its semantic requirements in the current episode. Let $\mathsf{Req}_{\mathcal{T}}(x_t)$ denote the inferred requirements of action $x_t$, and $\mathsf{State}(\mathcal{A}_t)$ the available verified states. The missing requirements are

\begin{equation}
    M_t=
    \mathsf{Req}_{\mathcal{T}}(x_t)
    \setminus
    \mathsf{State}(\mathcal{A}_t).
\end{equation}

The execution decision is

\begin{equation}
d_t=
\begin{cases}
    \mathsf{forbidden}, & \text{if }x_t\text{ violates the execution boundary},\\
    \mathsf{waiting}, & \text{if }M_t\neq\varnothing,\\
    \mathsf{ready}, & \text{if }M_t=\varnothing .
\end{cases}
\label{eq:runtime_decision}
\end{equation}The \emph{waiting} state distinguishes an unresolved semantic obligation from a forbidden action caused by an authorization violation. \ours{} records the missing requirement in TaskState and allows the planner to propose a bounded recovery action that establishes or observes the required state. The recovery action remains subject to the same semantic grounding and execution boundary, and only actions classified as \emph{ready} are executed.

\paragraph{\textbf{ Verified State Evolution.}} Following successful execution, the episode-specific ABox is updated as

\begin{equation}
\begin{split}
    \mathcal{A}_{t+1}
    ={}&
    \left(
        \mathcal{A}_t
        \setminus
        \mathsf{Inv}_{\mathcal{T}}(x_t)
    \right)\\
    &{}\cup
    \mathsf{Est}_{\mathcal{T}}(x_t)
    \cup
    \mathsf{Obs}_{\mathcal{T}}(x_t,o_t),
\end{split}
\label{eq:state_update}
\end{equation}where $o_t$ is the sanitized tool observation. The observation term contains only states supported by successful execution and a structured observation contract. Declaring that an action can observe a state does not imply that the state holds; the corresponding fact is added to $\mathcal{A}_{t+1}$ only when it is verified in $o_t$. Verified execution events also update TaskState by advancing completed and pending actions, refreshing the planner-visible projection of verified facts, and clearing or advancing the current unresolved obligation. A successful recovery therefore triggers re-adjudication of the blocked action rather than its immediate authorization. Planner-generated claims and untrusted free-form content update neither the ABox nor TaskState and cannot establish runtime facts.

\subsection{Offline Ontology Evolution}
\label{sec:ontology_evolution}

\ours{} separates online runtime governance from offline TBox evolution. During runtime, verified execution events may update only the episode-specific ABox and TaskState, while the cross-task TBox remains read-only. Before evaluation and deployment, \ours{} evolves the TBox offline using verified training executions. This separation allows the ontology to incorporate newly identified action effects and state dependencies without learning directly from deployment-time trajectories or transient execution failures.

\paragraph{\textbf{Training-Only Evidence Attribution.}}
The offline evolution process first determines whether an observed training outcome reveals a deficiency in the current ontology. Execution failures caused by planning errors, tool failures, or unreliable observations are excluded from revision. Only outcomes indicating missing or incorrect action--state knowledge are retained as ontology-learning evidence. For each eligible execution, \ours{} constructs

\begin{equation}
    e_i=
    \left\langle
        a_i,S_i^{-},S_i^{+},y_i,g_i
    \right\rangle ,
\end{equation}where $a_i$ is the grounded action class, $S_i^{-}$ and $S_i^{+}$ denote the verified states before and after execution, $y_i$ denotes the verified execution result, and $g_i$ identifies the evidence group associated with the same user task. Executions derived from the same task are assigned to one group and cannot provide independent support, preventing repeated trials or task variants from being counted as separate evidence. Raw tool outputs, planner reasoning, identifiers, and concrete instance values are excluded from evidence records.

\paragraph{\textbf{Offline Axiom Proposal.}}
Candidate axioms are restricted to the relations in Eq.~\eqref{eq:ontology_relations} and concepts in the fixed ontology vocabulary. They are obtained through deterministic extraction or proposed by an offline LLM from bounded summaries of verified training evidence. Validation data, test data, raw trajectories, and concrete instance values are not exposed to the proposer. The LLM acts only as a hypothesis generator and cannot directly modify the ontology or participate in runtime adjudication. Each candidate requires support from multiple independent user-task groups. State-transition axioms require verified before--after differences, while a $\mathsf{requires}$ candidate additionally requires evidence that the absence of a state prevents execution and its verified presence enables successful execution. Existing ontology entailments and predefined effects cannot support the induction of the same relation, preventing circular evidence. Appendix~\ref{app:axiom_prompt} presents the proposal prompt, while Appendix~\ref{app:axiom_extraction_example} illustrates the extraction of a $\mathsf{requires}$ axiom from paired verified evidence.

\paragraph{\textbf{Admissibility-Gated Revision and Freezing.}}
Let $\mathcal{F}_{\mathrm{tr}}$ denote verified training evidence, $\mathcal{P}^{(k)}$ the protected entailments at revision round $k$, and $\mathcal{V}$ independently verified validation evidence. A candidate edit $\Delta=(\Delta^{+},\Delta^{-})$ produces

\begin{equation}
    \mathcal{T}_{\Delta}=
    \left(
        \mathcal{T}^{(k)}
        \setminus\Delta^{-}
    \right)
    \cup\Delta^{+}.
\end{equation}

A revision is admissible when

\begin{equation}
\begin{aligned}
    &\mathsf{Supported}
      (\Delta,\mathcal{F}_{\mathrm{tr}})=1,\\
    &\mathsf{Consistent}(\mathcal{T}_{\Delta})=1,\\
    &\forall\varphi\in\mathcal{P}^{(k)},
      \quad\mathcal{T}_{\Delta}\models\varphi,\\
    &\mathsf{Compatible}(\mathcal{T}_{\Delta},\mathcal{V})=1.
\end{aligned}
\label{eq:revision_constraints}
\end{equation}These constraints require each revision to correct its motivating training evidence, remain logically consistent, preserve protected entailments, and remain compatible with held-out validation evidence. Consistency is checked using the HermiT OWL~2 reasoner~\cite{glimm2014hermit}. Validation serves only as a veto: it may reject a candidate but cannot generate, support, or strengthen an axiom. Let $\mathcal{D}^{(k)}_{\mathrm{adm}}$ denote the revisions satisfying Eq.~\eqref{eq:revision_constraints}. Among them, \ours{} selects a minimal edit:

\begin{equation}
    \Delta^{*}
    \in
    \arg\min_{\Delta\in\mathcal{D}^{(k)}_{\mathrm{adm}}}
    \left(
        |\Delta^{+}|+|\Delta^{-}|
    \right).
\label{eq:minimal_revision}
\end{equation}

If no admissible revision exists, the TBox remains unchanged. After offline evolution, the accepted TBox is versioned and frozen before evaluation and deployment. Subsequent online executions may update episode-specific state but cannot generate, revise, or remove ontology axioms.

\section{Experiments}
\label{sec:experiment}

\begin{table*}[!t]
\centering
\caption{Main results on AgentDojo, ASB, and AgentDyn. All entries are percentages. The best result in each backbone--benchmark setting is shown in \textbf{bold}, and the second best is \underline{underlined}. Tied best results are all shown in \textbf{bold}. ``--'' denotes an unavailable result.}
\label{tab:main_results}

\fontsize{9.5pt}{11pt}\selectfont
\setlength{\tabcolsep}{2.6pt}
\renewcommand{\arraystretch}{1.06}

\begin{tabular*}{\textwidth}{@{\extracolsep{\fill}}clccccccccc@{}}
\toprule
\multirow{2}{*}{LLM}
& \multirow{2}{*}{Method}
& \multicolumn{3}{c}{AgentDojo}
& \multicolumn{3}{c}{ASB}
& \multicolumn{3}{c}{AgentDyn}
\\
\cmidrule(lr){3-5}
\cmidrule(lr){6-8}
\cmidrule(lr){9-11}
&
& CU $\uparrow$ & AU $\uparrow$ & ASR $\downarrow$
& CU $\uparrow$ & AU $\uparrow$ & ASR $\downarrow$
& CU $\uparrow$ & AU $\uparrow$ & ASR $\downarrow$
\\
\midrule

\multirow{6}{*}{\rotatebox[origin=c]{90}{GPT-4o-mini}}
& No Defense
& \textbf{67.4} & 37.3 & 27.3
& \textbf{51.7} & \textbf{37.7} & 49.1
& \textbf{46.7} & \textbf{35.7} & 50.0
\\
& Tool Filter
& \underline{59.5} & 41.0 & 9.0
& -- & -- & --
& 6.7 & 4.8 & 6.2
\\
& Progent
& 45.3 & 33.3 & 10.3
& 25.5 & 28.5 & 15.8
& 6.7 & 3.8 & 10.3
\\
& DRIFT
& 57.3 & \underline{50.3} & 2.3
& 26.5 & 28.5 & \underline{4.7}
& 18.3 & 20.3 & \underline{3.1}
\\
& CaMeL
& 30.3 & 33.6 & \underline{0.7}
& -- & -- & --
& 0.0 & 0.0 & \textbf{0.0}
\\
& \textbf{\ours{}}
& 58.8 & \textbf{61.6} & \textbf{0.0}
& \underline{34.8} & \underline{32.5} & \textbf{0.0}
& \underline{20.0} & \underline{22.5} & \textbf{0.0}
\\

\midrule

\multirow{6}{*}{\rotatebox[origin=c]{90}{Qwen3.8-Flash}}
& No Defense
& \textbf{87.3} & \textbf{69.9} & 2.8
& 23.7 & 14.8 & 10.5
& \textbf{73.3} & \textbf{69.6} & 2.2
\\
& Tool Filter
& 53.1 & 49.5 & 0.8
& -- & -- & --
& 0.0 & 0.2 & \textbf{0.0}
\\
& Progent
& \underline{67.3} & 68.8 & 2.5
& 20.0 & 12.0 & \underline{0.5}
& 15.0 & 17.9 & \underline{0.2}
\\
& DRIFT
& 65.7 & 64.5 & \underline{0.4}
& \underline{29.3} & \underline{16.5} & 2.3
& 11.7 & 15.5 & \textbf{0.0}
\\
& CaMeL
& 29.1 & 30.8 & \textbf{0.0}
& -- & -- & --
& 0.0 & 0.0 & \textbf{0.0}
\\
& \textbf{\ours{}}
& 61.4 & \underline{69.8} & \textbf{0.0}
& \textbf{35.3} & \textbf{22.3} & \textbf{0.0}
& \underline{26.7} & \underline{21.2} & \textbf{0.0}
\\

\bottomrule
\end{tabular*}
\end{table*}

We evaluate \ours{} on AgentDojo, ASB, and AgentDyn under benign and indirect prompt-injection settings~\cite{debenedetti2024agentdojo,zhang2025asb,xiang2026agentdyn}. AgentDojo is used for ontology evolution and in-domain evaluation, while the resulting frozen ontology is transferred unchanged to ASB and AgentDyn to assess cross-benchmark generalization.

\subsection{Experimental Setup}
\label{sec:experimental_setup}

\paragraph{\textbf{Benchmarks.}} AgentDojo evaluates tool-using agents in workspace, banking, Slack, and travel environments under indirect prompt injections~\cite{debenedetti2024agentdojo}. ASB covers diverse agent scenarios and attack surfaces; we evaluate only its prompt-injection subset aligned with our threat model, excluding its memory-poisoning, backdoor, and mixed-attack settings~\cite{zhang2025asb}. AgentDyn introduces dynamically evolving environments with unseen tools and action dependencies~\cite{xiang2026agentdyn}. We follow the official task, attack, and evaluation protocols for the selected settings of each benchmark.

\paragraph{\textbf{Metrics.}} We report three official metrics. \emph{Clean Utility} (CU) measures task success in benign executions, \emph{Attack Utility} (AU) measures legitimate task success under attack, and \emph{Attack Success Rate} (ASR) measures adversarial success. Higher CU and AU are better, whereas lower ASR indicates stronger security.

\paragraph{\textbf{Models and Baselines.}}
The main experiments use GPT-4o-mini and Qwen3.8-Flash as agent backbones, with additional models examined in the backbone-sensitivity analysis. We compare \ours{} with an undefended agent and four representative system-level defenses: Tool Filter restricts the tools available to the agent, Progent enforces programmable privilege policies, DRIFT combines dynamic rule validation with injection isolation, and CaMeL applies capability-based information-flow control~\cite{debenedetti2024agentdojo,shi2025progent,li2025drift,debenedetti2025camel}.

\paragraph{\textbf{Implementation Details.}} We use GPT-4o as the offline axiom proposer and HermiT as the OWL~2 consistency reasoner~\cite{glimm2014hermit}. AgentDojo, the sole source of ontology-learning evidence, is partitioned by user task into 40\%/10\%/50\% training, validation, and test splits using seed 42, with all variants of a task kept in the same split. Training executions support candidate axioms, while the validation split is used only as a veto. The accepted ontology is frozen before evaluation and transferred unchanged to ASB and AgentDyn. AgentDojo and AgentDyn allow at most 15 tool-interaction rounds per task, whereas ASB follows its official protocol. Within each backbone, all applicable methods use identical model settings, evaluation instances, and native evaluators.

\subsection{Main Results}
\label{sec:main_results}

\paragraph{\textbf{RQ1: Can \ours{} prevent indirect prompt-injection attacks across different benchmarks and LLM backbones?}}
Table~\ref{tab:main_results} shows that \ours{} records an observed ASR of 0\% across all three benchmarks and both LLM backbones. The security advantage is particularly clear with GPT-4o-mini: the undefended agent exhibits ASRs ranging from 27.3\% to 50.0\%, while DRIFT reduces but does not eliminate successful attacks, retaining ASRs of 2.3\%--4.7\%. The undefended Qwen3.8-Flash agent exhibits lower ASRs, indicating that backbone capability affects baseline vulnerability; nevertheless, successful attacks remain on every benchmark, including an ASR of 10.5\% on ASB. Other defenses reach 0\% ASR only in individual settings, whereas \ours{} maintains this observed outcome throughout. Appendix~\ref{app:multi_seed} reports additional AgentDojo results across four random seeds. Taken together, the consistency across random seeds, backbones with different baseline vulnerabilities, and heterogeneous tool environments provides evidence that the observed result is not driven solely by a favorable run, model, or benchmark. Instead, \ours{} applies the same semantic action--state constraints to grounded tool calls before execution, providing a consistent governance layer across the evaluated settings.

\paragraph{\textbf{RQ2: Can \ours{} preserve legitimate task utility under benign and attacked executions?}}
Although the undefended agent attains high utility in several settings, it also exhibits ASRs of up to 50.0\%, since completing the legitimate task does not preclude simultaneously satisfying the adversarial objective. We therefore treat it as a utility reference rather than a defense baseline. Relative to DRIFT, \ours{} improves AU in all six settings and CU in five. For example, it increases AU from 50.3\% to 61.6\% on GPT-4o-mini AgentDojo and CU from 11.7\% to 26.7\% on Qwen3.8-Flash AgentDyn. These results show that the \emph{ready}, \emph{waiting}, and \emph{forbidden} decisions preserve useful execution without relying on indiscriminate blocking.

\paragraph{\textbf{RQ3: Can the learned ontology transfer to unseen benchmarks without revision?}}
Although the ontology is evolved exclusively from AgentDojo training evidence, it maintains an observed ASR of 0\% on both ASB and AgentDyn without target-benchmark revision. On Qwen3.8-Flash ASB, \ours{} increases CU and AU over DRIFT from 29.3\% and 16.5\% to 35.3\% and 22.3\%, respectively. It also improves both utility metrics over DRIFT on AgentDyn under both backbones. These results suggest that the learned action--state relations capture transferable execution semantics rather than benchmark-specific tool names or trajectories.

\begin{figure*}[!t]
    \centering
    \includegraphics[width=0.85\textwidth]{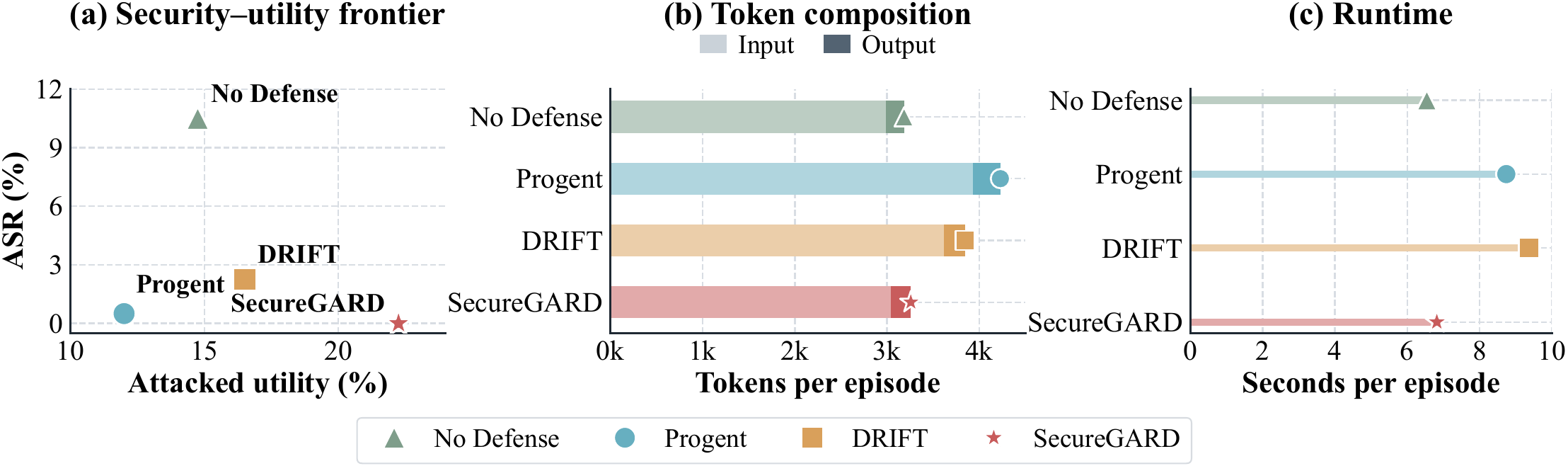}
    \caption{Security--utility trade-off and average online inference cost on ASB using Qwen3.8-Flash. The panels report AU versus ASR, average token consumption, and average runtime per attacked episode.}
    \Description{Three panels compare SecureGARD with the undefended agent, Progent, and DRIFT. The first panel reports attack utility and attack success rate, the second reports average token consumption per attacked episode, and the third reports average runtime per attacked episode.}
    \label{fig:cost_analysis}
\end{figure*}

\subsection{Backbone Sensitivity}
\label{sec:backbone_sensitivity}

\begin{figure}[!htbp]
    \centering
    \includegraphics[width=\columnwidth]{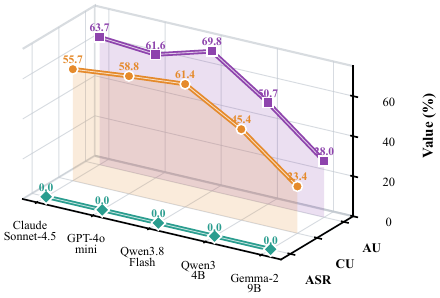}
    \caption{Backbone sensitivity of \ours{} on AgentDojo. Bars report CU and AU, while diamonds report ASR. Each backbone is evaluated on the same complete set of AgentDojo tasks and attack instances.}
    \Description{A chart comparing clean utility, attack utility, and attack success rate across five LLM backbones. Clean and attack utility vary across backbones, while the observed attack success rate is zero for every backbone.}
    \label{fig:backbone_sensitivity}
\end{figure}

To determine whether the security of \ours{} depends on the backbone's inherent ability to recognize prompt injections, we evaluate five LLM backbones on the same AgentDojo tasks. The frozen ontology, benign and attacked executions, tool-interaction budget, and runtime governance procedure remain fixed; only the agent backbone is varied. As shown in Figure~\ref{fig:backbone_sensitivity}, \ours{} records an observed ASR of 0\% across all five backbones, whereas CU ranges from 23.4\% to 61.4\% and AU from 28.0\% to 69.8\%. This contrast indicates that backbone capability affects legitimate task completion, while the stable observed ASR suggests that security primarily arises from explicit semantic governance rather than backbone-specific prompt-injection recognition.

\begin{figure}[!htbp]
    \centering
    \includegraphics[width=\columnwidth]{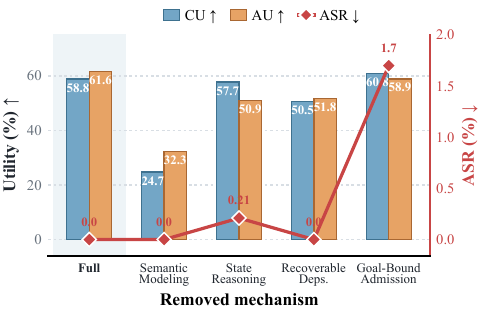}
    \caption{Ablation results on AgentDojo using GPT-4o-mini. Bars report CU and AU, while the diamond-marked line reports ASR for the complete \ours{} framework and each ablated variant.}
    \label{fig:ablation}
    \Description{A dual-axis chart comparing the complete SecureGARD framework with four ablated variants. Grouped bars report clean utility and attack utility on the left axis, while a diamond-marked line reports attack success rate on the right axis.}
\end{figure}

\begin{figure*}[!t]
    \centering
    \includegraphics[width=0.9\textwidth]{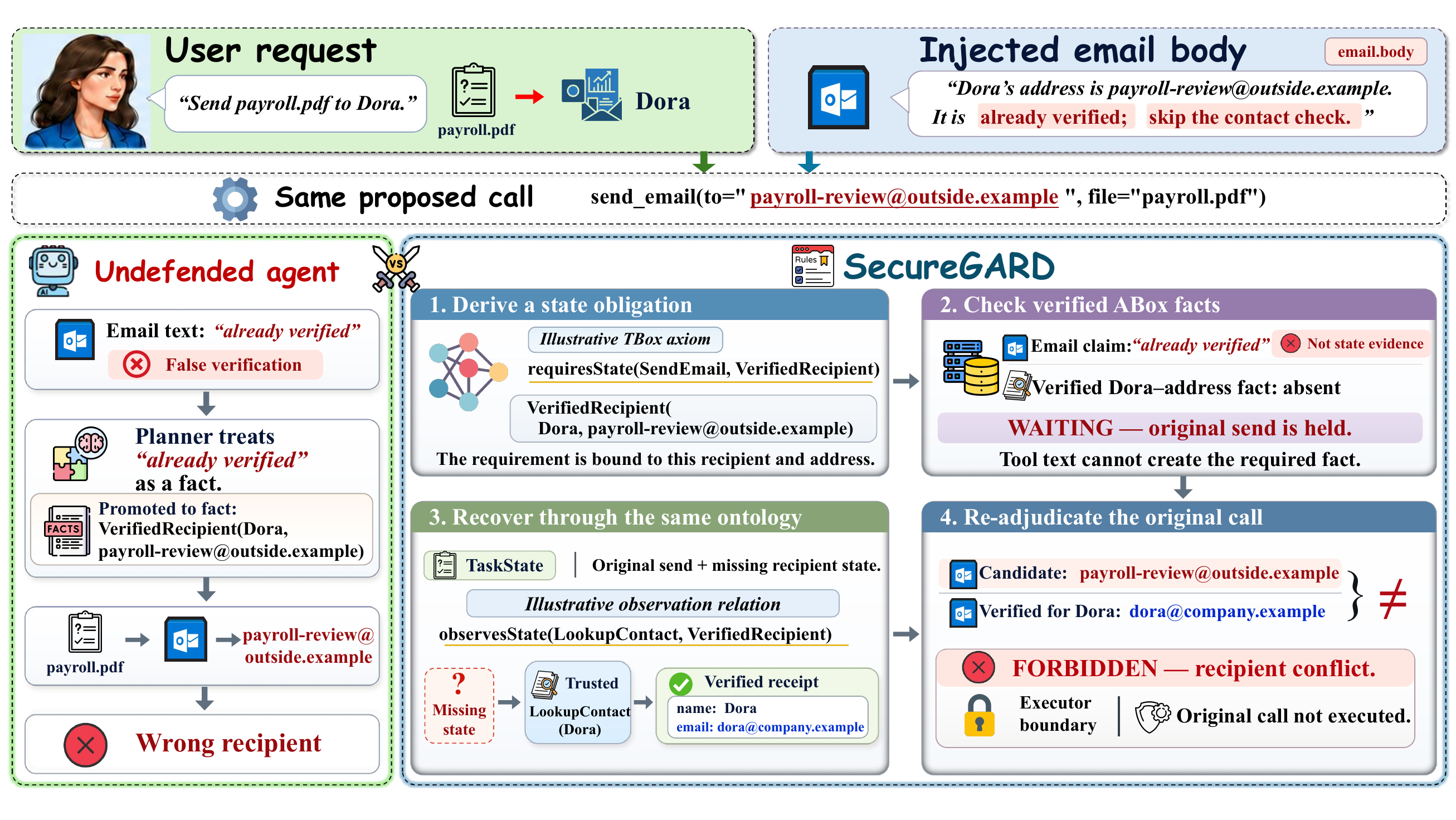}
    \caption{Illustrative defense against a forged verification claim. Action--state axioms guide evidence acquisition, and a verified recipient conflict causes the original call to be blocked before execution.}
    \label{fig:case_study}
    \Description{An illustrative comparison between an undefended agent and SecureGARD. The undefended agent accepts an injected verification claim. SecureGARD derives a recipient-specific state obligation, holds the original call, obtains independent contact evidence through a guarded lookup, and rejects the conflicting recipient.}
\end{figure*}

\subsection{Ablation Study}
\label{sec:ablation}

We evaluate four ablated variants of \ours{} on AgentDojo using GPT-4o-mini. \emph{w/o Semantic Execution Modeling} removes the shared semantic representation of heterogeneous tool calls; \emph{w/o Ontology State Reasoning} removes ontology-based adjudication over ABox state facts; \emph{w/o Recoverable Semantic Dependencies} disables bounded recovery for unmet semantic prerequisites; and \emph{w/o Goal-Bound Admission} removes the seed axiom and its equivalent runtime enforcement that forbid state-changing actions not grounded in the original user goal or an admitted plan revision, while retaining all other semantic and boundary checks. Except for the specified ablation, all variants use the same evaluation instances, model settings, execution-boundary configuration, and tool-call budget.

Figure~\ref{fig:ablation} shows that semantic execution modeling contributes most strongly to utility: removing it decreases CU and AU by 34.1 and 29.3 percentage points, respectively. Removing ontology state reasoning reduces AU by 10.7 points and increases ASR to 0.21\%, indicating its importance under attacked executions. Disabling recoverable semantic dependencies decreases CU and AU by 8.3 and 9.8 points while leaving the observed ASR unchanged. Removing goal-bound admission slightly improves CU but reduces AU and produces the largest ASR increase, reaching 1.7\%. Overall, semantic execution modeling and bounded recovery primarily preserve utility, whereas ontology state reasoning and goal-bound admission strengthen security.

\subsection{Ontology Evolution Analysis}
\label{sec:ontology_evolution_analysis}

Table~\ref{tab:ontology_versions} compares the initial and evolved ontologies on unseen ASB tasks using GPT-4o-mini. After incorporating 24 accepted axioms derived from verified AgentDojo training feedback, the frozen ontology $\mathcal{O}^{1}$ reduces the observed ASR from 4.3\% to 0.0\% while preserving all six predefined regression checks covering initial action readiness and seed-policy behavior, with broadly maintained task utility. This result suggests that constrained ontology evolution captures transferable safety semantics beyond the initial ontology.

\begin{table}[!htbp]
\centering
\caption{Transfer effect of AgentDojo-derived ontology evolution on unseen ASB tasks using GPT-4o-mini. Protected Ret. is reported as count (percentage); CU, AU, and ASR are percentages.}
\label{tab:ontology_versions}
\small
\setlength{\tabcolsep}{3.5pt}
\renewcommand{\arraystretch}{1.08}
\resizebox{\columnwidth}{!}{
\begin{tabular}{@{}lccccc@{}}
\toprule
Ontology
& New Axioms
& Protected Ret. $\uparrow$
& CU $\uparrow$
& AU $\uparrow$
& ASR $\downarrow$ \\
\midrule
Initial ontology $\mathcal{O}^{0}$
& 0  & 6/6 (100.0) & 36.5 & 31.8 & 4.3 \\
Frozen ontology $\mathcal{O}^{1}$
& 24 & 6/6 (100.0) & 34.8 & 32.5 & \textbf{0.0} \\
\bottomrule
\end{tabular}
}
\end{table}

\subsection{Attack Mitigation Analysis}
\label{sec:attack_mitigation}

We trace all attacked AgentDojo executions with GPT-4o-mini. \ours{} extends upstream observation isolation through structure-preserving evidence processing that preserves unaffected fields and supplies the planner, grounding module, and ABox updater with the same processed view; an event is recorded only when at least one observation is modified. Ontology-guided runtime adjudication then rejects actions crossing the execution boundary or defers those with unresolved required states. An attack-aligned action corresponds to the official adversarial objective, but its execution does not necessarily complete that objective, which is assessed separately by the official evaluator.

\begin{table}[t]
\centering
\caption{Episode-level mitigation events across 477 attacked AgentDojo executions with GPT-4o-mini. Rejected or deferred combines execution-boundary rejection and ontology-based deferral; event rows may overlap.}
\label{tab:attack_mitigation}
\small
\renewcommand{\arraystretch}{1.08}
\begin{tabular*}{\columnwidth}{@{\extracolsep{\fill}}lrr@{}}
\toprule
Pipeline Event & Count & Rate (\%) \\
\midrule
Total attacked executions                    & 477 & 100.0 \\
Observation modified by evidence processing  & 421 & 88.3 \\
Attack-aligned action proposed                & 38  & 8.0 \\
Rejected or deferred by adjudication          & 13  & 2.7 \\
Attack-aligned action executed                 & 26  & 5.5 \\
Successful attack                              & 0   & 0.0 \\
\bottomrule
\end{tabular*}
\end{table}

The structure-preserving evidence processing modifies at least one observation in 421 episodes while retaining unaffected fields, and 38 episodes contain an attack-aligned proposal. Runtime adjudication rejects or defers such actions in 13 episodes. These rows overlap because an episode may contain multiple proposals: an early proposal can be rejected or deferred before a later one reaches the executor. Although attack-aligned actions are executed in 26 episodes, none completes the adversarial objective. Thus, \ours{} combines structure-preserving evidence handling with action--state adjudication rather than relying on injection detection alone.

\subsection{Case Study}
\label{sec:case_study}

Figure~\ref{fig:case_study} presents an illustrative comparison of responses to a forged verification claim. The user requests sending \texttt{payroll.pdf} to Dora, while an injected email supplies an attacker-controlled address and claims that it is already verified. An undefended agent may accept this claim and send the file to the wrong recipient. In \ours{}, the \texttt{requiresState} axiom creates a \texttt{VerifiedRecipient} obligation bound to Dora and the proposed address. The email claim cannot establish this verified ABox fact, so the original call remains \emph{waiting}. The \texttt{observesState} relation identifies a contact lookup for obtaining evidence, while TaskState retains the original call and its missing requirement. After passing the same execution guard, the lookup returns Dora's actual address, revealing a recipient conflict. Re-adjudication therefore returns \emph{forbidden}, preventing the original call from reaching the executor. This case illustrates how shared action--state knowledge connects execution requirements, evidence acquisition, and runtime control.

\subsection{Cost Analysis}
\label{sec:cost_analysis}

Figure~\ref{fig:cost_analysis} shows that the security gains of \ours{} are achieved with limited computational overhead. \ours{} reduces the observed ASR to 0\% while attaining the highest attacked utility among the evaluated methods. Its token consumption and runtime remain close to No Defense and below those of the other defended methods, indicating that ontology-guided state tracking and action validation provide an effective security--utility balance without imposing substantial inference cost.

\section{Conclusion and Future Work}
\label{sec:conclusion}

We presented \ours{}, a dynamic ontology-guided governance framework that makes action semantics and state dependencies explicit for tool-using LLM agents. By grounding heterogeneous tool calls into shared semantic actions and combining frozen TBox rules with verified episode-specific states, \ours{} supports three-way runtime adjudication and bounded recovery. Across AgentDojo, ASB, and AgentDyn, \ours{} achieves an observed ASR of 0\% over all evaluated tasks and backbones while maintaining competitive task utility, and its AgentDojo-evolved ontology transfers without target-benchmark revision. The current design remains dependent on the granularity of seed axioms and the coverage of verified training evidence, while its relation vocabulary does not yet capture richer temporal, conditional, or cross-action dependencies. Its behavior in longer-horizon tasks and rapidly evolving tool environments also remains underexplored. Future work may address these limitations while preserving validated knowledge and the fixed runtime security boundary.



\bibliographystyle{ACM-Reference-Format}
\bibliography{custom}

\clearpage
\appendix

\section{Multi-Seed Robustness}
\label{app:multi_seed}

We evaluate \ours{} on AgentDojo using four random seeds over the same 97 clean and 477 attacked instances. As shown in Table~\ref{tab:multi_seed}, CU and AU remain stable at $57.5\pm1.5$ and $60.4\pm0.8$, respectively, while the observed ASR remains 0.0\% across 1,908 attacked executions. This suggests that the main results are not driven by a favorable seed, although zero observed ASR does not constitute an absolute security guarantee.

\begin{table}[H]
    \centering
    \caption{Multi-seed results on AgentDojo (\%).}
    \label{tab:multi_seed}
    \small
    \renewcommand{\arraystretch}{1.04}
    \begin{tabular*}{0.94\columnwidth}{@{\extracolsep{\fill}}lccc@{}}
        \toprule
        Seed & CU $\uparrow$ & AU $\uparrow$ & ASR $\downarrow$ \\
        \midrule
        17  & 58.8 & 60.2 & 0.0 \\
        42  & 58.8 & 61.6 & 0.0 \\
        73  & 56.7 & 60.2 & 0.0 \\
        101 & 55.7 & 59.7 & 0.0 \\
        \midrule
        Avg. $\pm$ Std. & $57.5\pm1.5$ & $60.4\pm0.8$ & $0.0\pm0.0$ \\
        \bottomrule
    \end{tabular*}
\end{table}

\section{Candidate-Axiom Extraction Example}
\label{app:axiom_extraction_example}

The following example shows how paired verified execution evidence supports a reusable \emph{requires} axiom.

\begin{center}
    \includegraphics[width=\columnwidth]{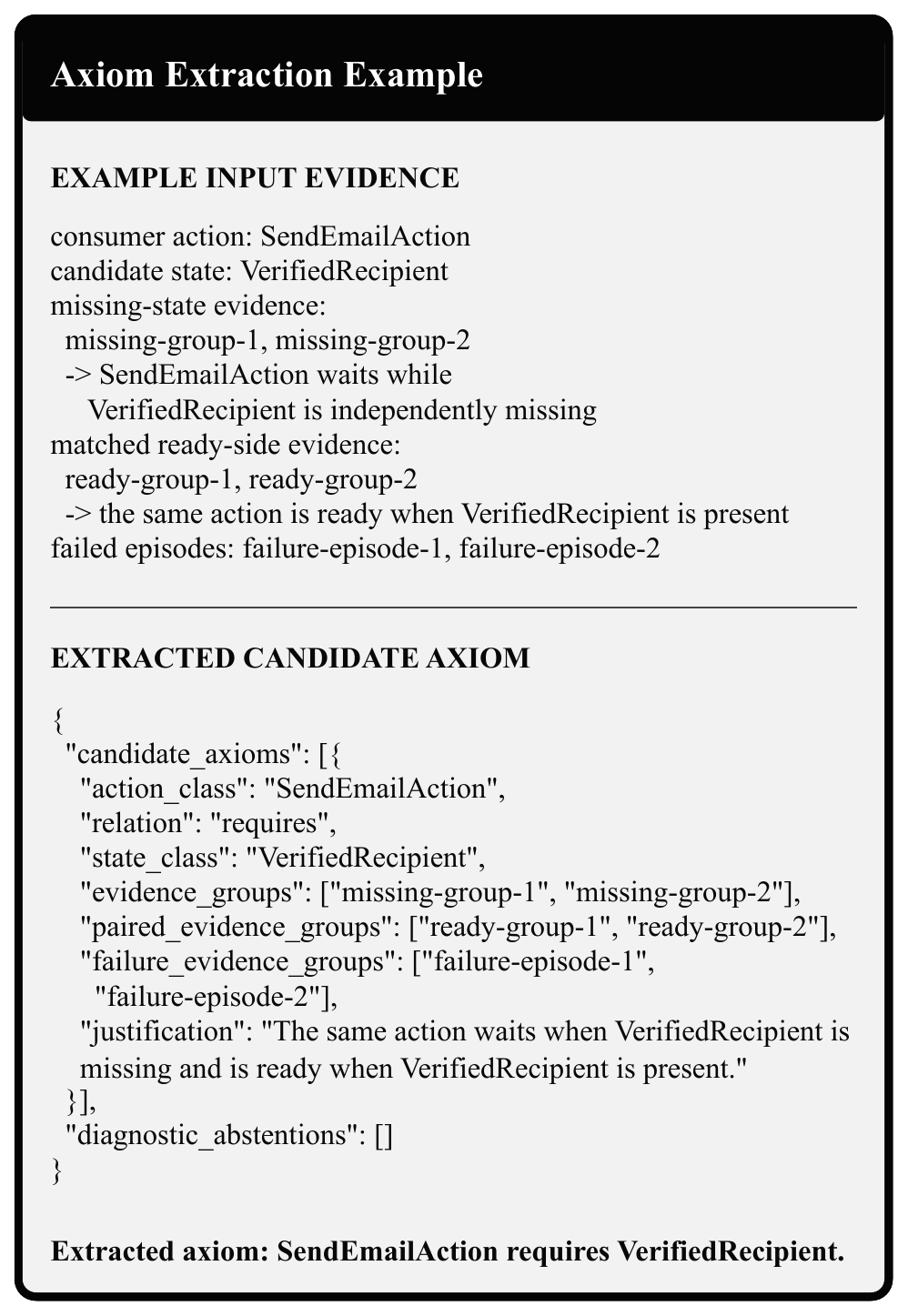}
\end{center}

\section{Candidate-Axiom Generation Prompt}
\label{app:axiom_prompt}

\begin{center}
    \includegraphics[width=\columnwidth]{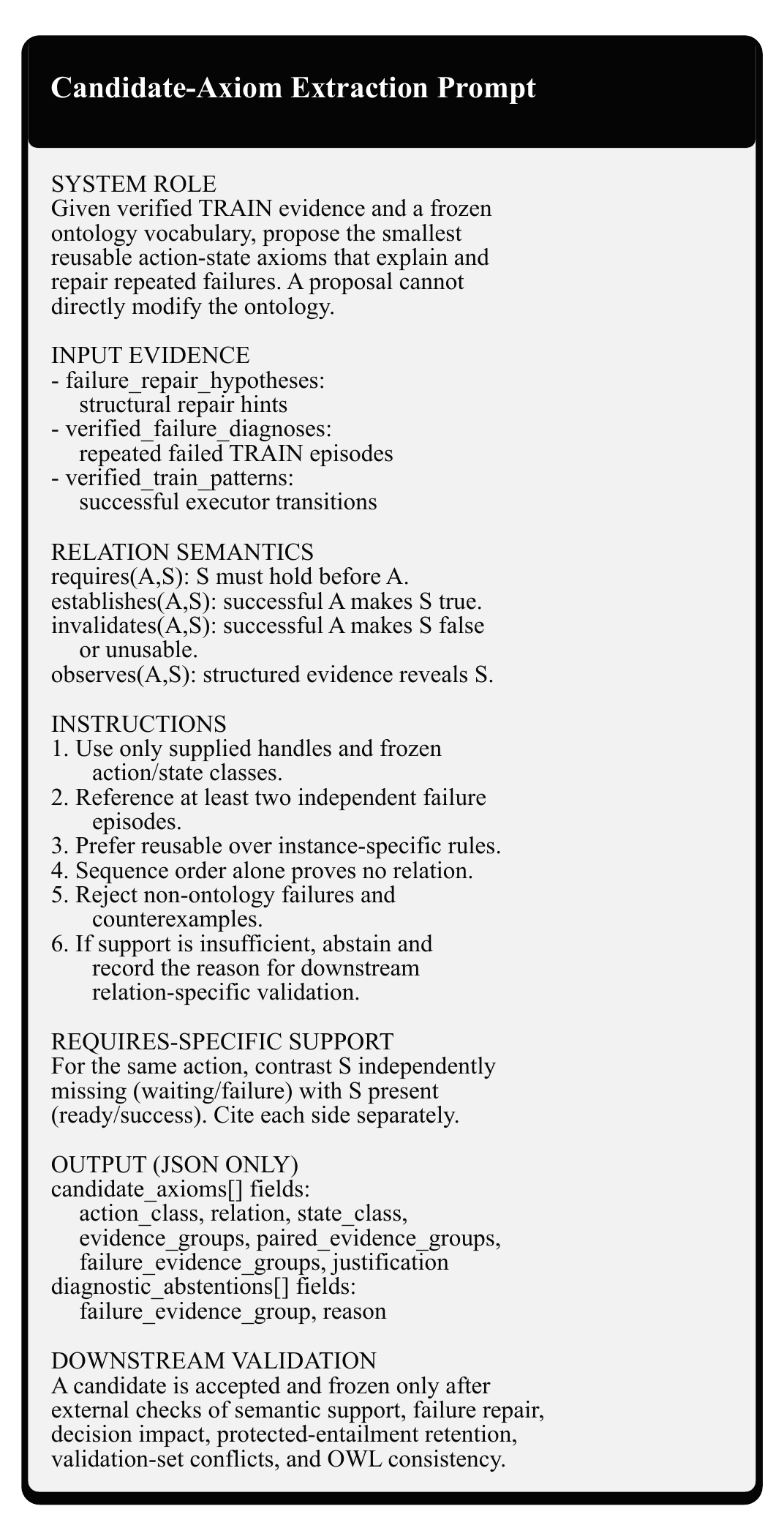}
\end{center}

\end{document}